\documentclass{article}

\usepackage{arxiv}

\usepackage[utf8]{inputenc} 
\usepackage[T1]{fontenc}    
\usepackage{hyperref}       
\usepackage{url}            
\usepackage{booktabs}       
\usepackage{amsfonts}       
\usepackage{nicefrac}       
\usepackage{microtype}      
\usepackage{lipsum}		
\usepackage{graphicx}
\usepackage{natbib}
\usepackage{doi}
\usepackage{amsmath}
\usepackage{graphics}
\usepackage{caption}
\usepackage[caption=false,font=normalsize,labelfont=sf,textfont=sf]{subfig}
\usepackage{multirow}
\usepackage{subfig}
\usepackage{hyperref}

\title{Developing a Tutoring Dialog Dataset \\ to Optimize LLMs for Educational Use}

\author{Menna Fateen \\
Graduate School of Information Science and Electrical Engineering\\
Kyushu University\\
Fukuoka, 8190395, Japan \\
\texttt{menna.fateen@m.ait.kyushu-u.ac.jp} \\
\And
{Tsunenori Mine} \\
Faculty of Information Science and Electrical Engineering\\
Kyushu University\\
Fukuoka, 8190395, Japan \\
\texttt{mine@ait.kyushu-u.ac.jp } \\
}

\renewcommand{\shorttitle}{\textit{arXiv} Template}

\hypersetup{
pdftitle={A template for the arxiv style},
pdfsubject={q-bio.NC, q-bio.QM},
pdfauthor={David S.~Hippocampus, Elias D.~Striatum},
pdfkeywords={automatic short answer scoring, large language models, retrieval-augmented generation},
}

\begin{document}
\maketitle

\begin{abstract}
  Recent advances in large language models (LLMs) have shown promise for scalable educational applications, but their use in dialog-based tutoring systems remains challenging due to the need for effective pedagogical strategies and the high costs associated with expert-curated datasets. Our study explores the use of smaller, more affordable LLMs for one-on-one tutoring in the context of solving reading comprehension problems. We developed a synthetic tutoring dialog dataset, evaluated by human teachers, and fine-tuned a smaller LLM using this dataset. Furthermore, we conducted an interactive experiment comparing the performance of the fine-tuned model with a larger model in real-world tutoring scenarios. Our results show that the fine-tuned model performs on par with the larger model but at a lower cost, demonstrating a viable, cost-effective approach for implementing LLM-based tutoring systems in educational settings.

\end{abstract}


\section{Introduction}
 Large language models have become versatile tools, serving as personal assistants with various capabilities such as question answering, language translation, etc \cite{kaddour2023challenges}.  They have proven to be effective in many task-oriented dialog systems where they provide answers to complex questions \cite{hudevcek2023large}. 

However, in educational settings, there is a need for more than just delivering accurate answers. Effective pedagogical strategies are essential for enhancing the educational value of LLMs. Socratic  questioning \cite{lepper2002wisdom}, for instance, helps deepen understanding and encourage critical thinking by prompting learners to reflect more on their answers. Encouraging a growth mindset, which emphasizes that intelligence and abilities can be developed through effort and learning strategies \cite{dweck2006mindset}, can help students face challenges positively and learn from their errors. This approach has not only immediate benefits but also long-term effects, as studies have shown that growth mindset language positively influences students’ academic performance over time \cite{hecht2023efficiently}. Showing care and empathy makes learners more engaged and confident, creating a supportive atmosphere that helps them learn and achieve better \cite{roorda2011influence,pianta2017teacher,robinson2022framework}.  These requirements are necessary for an educational tutor and differ from those typically expected of personal assistants, who prioritize delivering accurate and immediate answers. An effective educational tool must guide learners through the educational process rather than focusing solely on results. 



To implement these strategies in an LLM-driven educational setting, we propose the development of a tutoring dialog dataset. This dataset aims to simulate realistic student-tutor interactions that are not only linguistically accurate but also pedagogically effective. It will allow us to test the capability of LLMs to function as educational tutors. We explore the feasibility of using Transformer-based mixture of experts LLMs, specifically Mistral `8x7b' and `7b' models as one-on-one tutors. Our focus is on reading comprehension tasks, where students read a passage and answer inference questions—a task particularly challenging for young learners. 

Our research aims to address the following research questions:
\begin{itemize}
    \item \textbf{RQ1:} Can we develop a tutoring dialog dataset that effectively simulates realistic student-tutor interactions in terms of pedagogical effectiveness and linguistic correctness?
    \item \textbf{RQ2:} Can this dataset be used to fine-tune smaller models to achieve learning outcomes comparable to those delivered by larger models?

\end{itemize}

The creation of this dataset serves a dual purpose: (a) it tests the capability of LLMs to function as tutors and (b) provides a controlled environment to evaluate and refine their performance. This approach aligns with the need for scalable, cost-effective educational solutions, particularly in resource-limited settings. Hence, our second research question (RQ2) explores the viability of fine-tuning smaller, more economical LLMs with our dataset. By comparing the performance of a smaller fine-tuned model to its larger, more resource-intensive counterpart, we aim to demonstrate that smaller models can achieve comparable outcomes, thus enhancing their practicality and accessibility for widespread educational use.


Human expert evaluations first confirm the dataset’s effectiveness in capturing realistic tutoring dynamics and incorporating essential pedagogical strategies. Our findings also indicate that the fine-tuned smaller model performs comparably in key educational metrics, offering a cost-effective and scalable alternative for educational settings. However, it is crucial to recognize that LLMs especially smaller models may exhibit limitations such as repetitiveness and hallucinations. Acknowledging these challenges is essential for future research aimed at enhancing the reliability and educational integrity of LLM applications.

\begin{figure}[]
  \centering 
  \includegraphics[width=0.5\columnwidth]{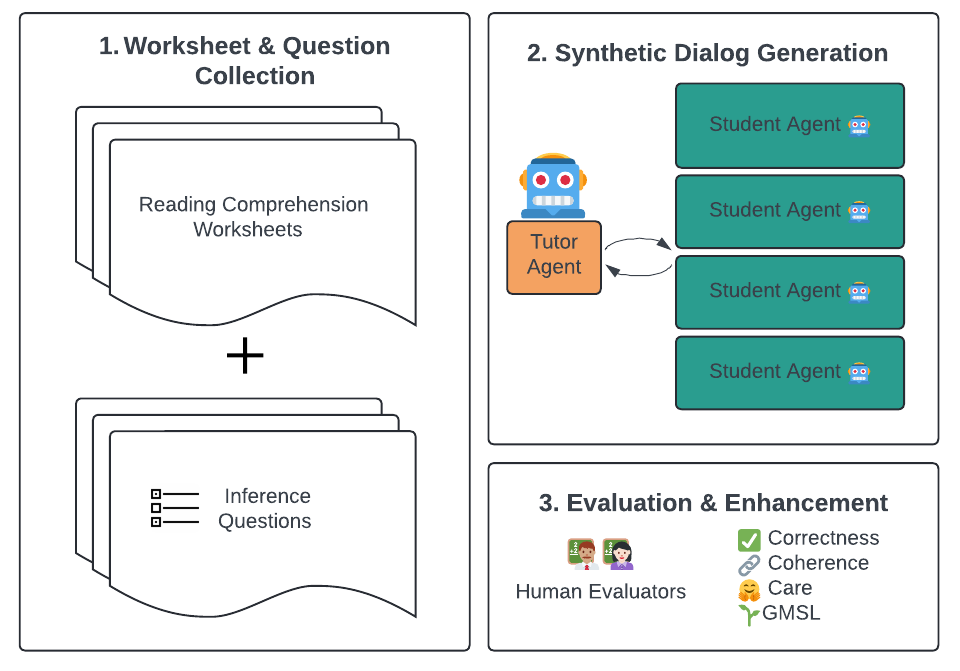} 
  \caption{Dataset construction method.}  
  \label{fig:method} 
\end{figure}

In summary, the contributions of this research are as follows:
\\
\\
\begin{itemize}
    \item \textbf{Creation of a specialized dataset}: We developed a tutoring dialog dataset that simulates realistic student-tutor interactions for reading comprehension tasks, incorporating essential pedagogical strategies. Figure \ref{fig:method} illustrates the dataset construction process.
    \item \textbf{Performance enhancement through fine-tuning with the dataset}: By fine-tuning a smaller-scale LLM using the newly created dataset, we achieved performance comparable to that of a larger-scale LLM. Notably, the fine-tuned model demonstrated enhanced performance in key educational aspects i.e. Helpfulness, Care, and Growth Mindset Language (GMSL).
    \item \textbf{Empirical evaluation with educational stakeholders}: We conducted  evaluations of the dialog performance of the LLMs with actual teachers and learners, providing a comprehensive assessment of the models' effectiveness in real-world educational settings.
\end{itemize}
We make our dataset and code publicly available\footnote{Dataset available at \url{http://github.com/} \textit{redacted for review}}

\section{Related Work}
Educational dialogue systems have transformed from simple rule-based models to sophisticated AI-driven tutors capable of engaging students in meaningful interactions \cite{wollny2021we}. 

With the emergence of powerful language models like GPT-3 and BART, their application in educational dialogue systems has expanded significantly \cite{macina2023opportunities}. For instance, \cite{tack2022ai} tested the pedagogical abilities of generative models such as Blender \cite{roller2020recipes} and GPT-3 and found that despite good conversational uptake, these models significantly underperform compared to human teachers in key educational aspects, notably in helpfulness.

\subsection{Tutoring Dialog Datasets}

Building educational chatbots requires large amounts of high-quality expert-curated training data. For example, \cite{wang2023step} presented a dataset of real tutoring conversations, annotated by experts with their decision making process in error remediation. However, such datasets are scarce and can be expensive and time-consuming to collect. To address this challenge, many approaches have been proposed. For instance, \cite{suresh2022talkmoves} developed a dataset of teacher-student interactions in math classrooms, which was used to train a model to classify teacher talk moves. Another dataset is built via crowdworkers by roleplaying both the teacher and the student in a language translation task \cite{stasaski-etal-2020-cima}. More recently, \cite{macina2023mathdial} paired teachers with LLMs simulating students to solve GSM8K problems \cite{cobbe2021training}. However, these methods are limited in scalability and are often domain-specific, making them less suitable for general educational applications.

\subsection{Synthetic Dialog Datasets}

Synthetic data generation has been explored as a cost-effective alternative to collecting large amounts of real-world data. \cite{chen2023places} used few-shot prompting to generate a social conversation dataset, while \cite{zhan2023socialdial} used GPT-3 to generate a socially-aware dialog dataset based on Chinese social culture. \cite{kim2023soda} contextualized social commonsense knowledge from a knowledge graph to generate an open-domain dialog dataset. However, these studies focus on social dialogues and lack the pedagogical strategies required for educational dialogues. A notable exception is \cite{wang2024book2dial}, which developed a synthetic teacher-student interaction dataset grounded in textbook content. However, in their model, the teacher directly provides students with answers since their focus is on information exchange. 

\section{Methodology}

\subsection{Worksheet and Question Collection}

In order to ground our tutor agent on real-world educational contexts, we collected 23 reading comprehension worksheets from various educational websites. Each worksheet includes a passage and a set of inference questions related to the passage. The questions are designed to assess the student's understanding of the text and require them to draw conclusions based on the information provided in the passage. All 63 questions have multiple-choice answers, with one correct answer and three distractors. The worksheets were selected to cover a wide range of topics and difficulty levels to ensure the diversity of the dataset. Grade levels of the worksheets range from 2nd to 5th grade and span both fiction and non-fiction passages. With an average length of 208 words, the paragraphs range in length from 24 to 420 words. Question types include: context clues for vocabulary, inferring sequence of events, drawing conclusions, and making predictions.

\subsection{LLM Selection}
In both the development of the dataset and the interactive tutoring experiment, we selected the Mistral 7b and Mixtral8x7b models \cite{jiang2023mistral,jiang2024mixtral}, for their cost-effectiveness and high performance capabilities. As open-source tools, these models offer a significant cost advantage by eliminating licensing fees, making them ideal for widespread implementation and facilitating broader academic and practical applications. 
\subsection{Dialog Tutoring Session}

\begin{figure}[h!]
  \centering 
  \includegraphics[width=0.5\columnwidth]{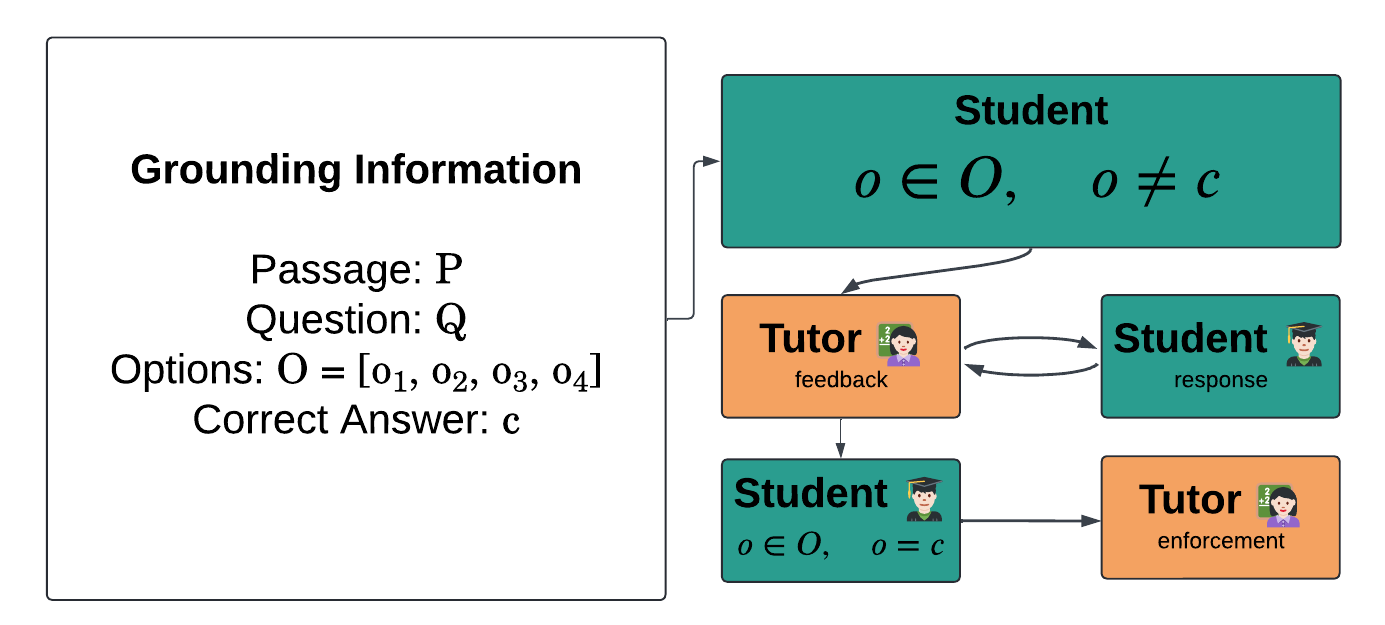} 
  \caption{Dialog tutoring generation process.}  
  \label{fig:gen} 
\end{figure}

The dialog tutoring generation process involves a series of interactions between two participants: a student $s$ and a tutor $T$. We view each tutoring session as a sequence of turns, with each turn consisting of a series of tokens from a vocabulary set $V$.
Each session is initiated by a passage $P$ and a multiple-choice question $Q$ that has 4 options $O = \{o_1, o_2, o_3, o_4\}$. Only one option, $o_i$ ($i \in{1,..,4}$) is the correct answer $c$. To mimic a real tutoring scenario, we initially have the student model choose an incorrect option $o \neq c$. This setup helps simulate a situation where the student needs help to find the right answer. The flow of the dialog tutoring generation process is shown in Figure \ref{fig:gen}.

\subsection{Dialog Tutoring Generation}

We implemented the tutor agent using the Mixtral 8x7b model \cite{jiang2024mixtral}. The tutor agent's responses are generated using zero-shot prompt tuning. We repeatedly adjusted our prompts until the tutor's performance was satisfactory. Our approach to creating prompts follows the `Flipped Interaction Pattern' \cite{white2023prompt}. In this method, we first specify the goal of the LLM, then set the context for how long the session should last, and finally limit the length of the generated response. The tutor's main aim is to help the student find the correct answer while encouraging them to think more deeply and develop a growth mindset. To keep responses concise, we instruct the tutor to limit replies to a maximum of three sentences.

The final tutor prompt we used was: \textit{`As an English tutor, your role is to help young learners improve their reading and understanding skills. The student has chosen an incorrect answer. Your goal is to guide them to find the correct answer, by thinking deeply and looking for clues in the text. When they get it right, say: 'Exactly! That's the right answer. You can now close this tab and continue with the rest of your worksheet.' Use easy words and encourage them to ask questions and think creatively. Help them connect the dots without giving the answer directly. Remember, short sentences and clear hints are key.`}

To mimic student-tutor interactions, we also used the Mixtral 8x7b model to simulate different student agents.
We designed four distinct learner profiles that represent different learning styles and details on these profiles are available in the appendix. Each student agent is provided with the passage, the question, its corresponding options, and the previous chat history as grounding information, but the correct answer is not given. The student agent first responds to the tutor's feedback on their initial incorrect choice. We set the student LLM to generate only one sentence at a time, to mimic a typical short student response in a chatbot interaction. The tutor then provides the next piece of guidance, and the dialogue continues until the student identifies the correct answer. To finish the dialog session, we wait until the correct answer is reached and the tutor instructs the student to \textit{`close the tab`} and move on to the next part of the worksheet or until the maximum number of turns (10) is reached.

\subsection{Dataset Analysis}
\subsubsection*{Tutor Turns}

Throughout the development of our dialog generation process, we continually analyzed each iteration to refine and enhance our approach. In this section, we present a comprehensive analysis of the final version, focusing on both the quantitative and qualitative metrics.

We begin by examining quantitative metrics such as the number of conversational turns and the talktime, measured by the number of words for both tutor and student agents. The distribution of talktime for the tutor and student agents is depicted in Figure \ref{fig:talktime}. As illustrated in the figure, the tutor agent's responses are consistently longer than those of the student agent.

To assess the quality of tutors' responses, we employed a BERT model finetuned on tutoring data \cite{wang2024convokit} to analyze conversational uptake, assessing whether tutors build upon students' contributions. The results of this analysis for the dataset are detailed in the middle of Table \ref{tab:analysis-res}. Additionally, we utilized a finetuned RoBERTa classifier to evaluate the presence of focusing questions \cite{alic2022computationally}—questions that engage students in articulating their thoughts and reasoning. The frequency of these questions within our dataset is shown in the top of Table \ref{tab:analysis-res}.


  \begin{table}[]
    \centering
    \caption{Classification analysis of generated tutor and student responses. Uptake and Focusing are tutor metrics, while Reasoning, No Talk, Claim, and Evidence are student metrics. Numbers represent the percentage of responses classified as each category.}
    \begin{tabular}{l|c}
    \hline
    \multicolumn{2}{c}{\textbf{Turns}} \\ \hline
    Average Turns & 9.6 \\
    Maximum Turns & 13 \\
    Minimum Turns & 4 \\ \hline
    \multicolumn{2}{c}{\textbf{Tutor}} \\ \hline
    Uptake & 0.86 \\
    Focusing & 0.28 \\ \hline
    \multicolumn{2}{c}{\textbf{Student}} \\ \hline
    Reasoning & 0.15 \\
    Talk Move 0: No Talk & 0.55 \\
    Talk Move 3: Claim & 0.29 \\
    Talk Move 4: Evidence & 0.01 \\ 
    \hline
    \end{tabular}
    \label{tab:analysis-res}
\end{table}

Finally, the types of tutor responses are classified with the TalkMoves \cite{suresh2022talkmoves} finetuned ELECTRA model. Due to the subject-specific nuances between reading comprehension and the original training data based on math classrooms, we found that 44\% of the tutor interactions were classified as `Other'. The remaining 56\% were classified as `Keeping Everyone Together' which encompasses actions such as posing yes/no questions, managing turns, getting students' attention, keeping students on task, and managing the classroom. Qualitative examples of positive and negative uptake, talk moves, and focusing questions can be found in the appendix.

\subsubsection*{Student Turns}

While the primary focus was on the tutor's responses, significant effort was also made to realistically simulate student behavior. This involved classifying student talk moves \cite{suresh2022talkmoves} using a finetuned ELECTRA classifier for 5 classes: 0: No Talk Move Detected, 1: Relating to Another Student, 2: Asking for More Information, 3: Making a Claim, 4: Providing Evidence or Reasoning. The distribution of these classes in the dataset is shown in the bottom of Table \ref{tab:analysis-res}. In addition, we examined student reasoning using a RoBERTa classifier that was finetuned on the NCTE math classroom transcripts \cite{demszky2023ncte}. However, as indicated in Table \ref{tab:analysis-res}, a low percentage of student responses were classified as `Reasoning' (15\%), reflecting the task's requirement for students to infer information rather than provide mathematical explanations.
\begin{figure}[h!]
  \centering 
  \includegraphics[width=0.5\columnwidth]{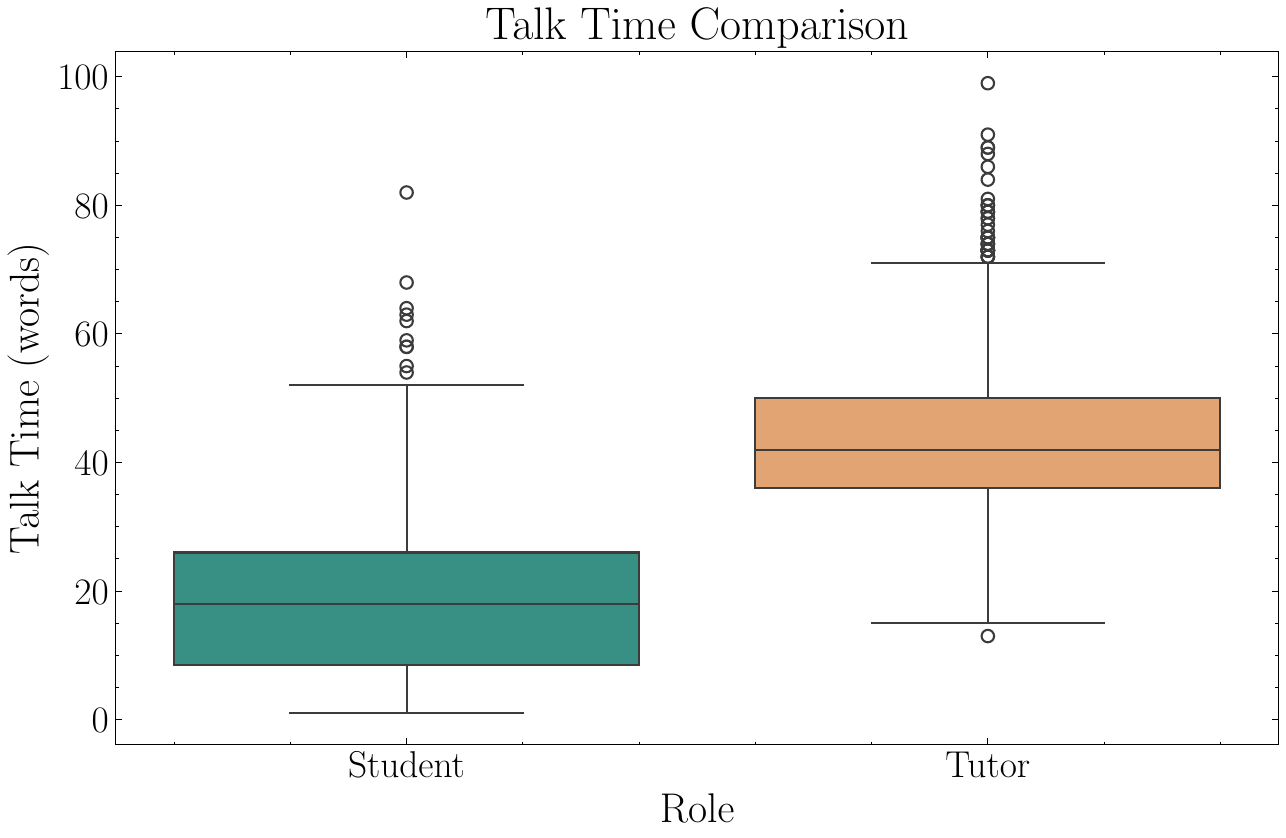} 
  \caption{Talktime distribution for tutor and student agents.}  
  \label{fig:talktime} 
\end{figure}

\begin{figure}[h!]
  \centering 
  \includegraphics[width=0.5\columnwidth]{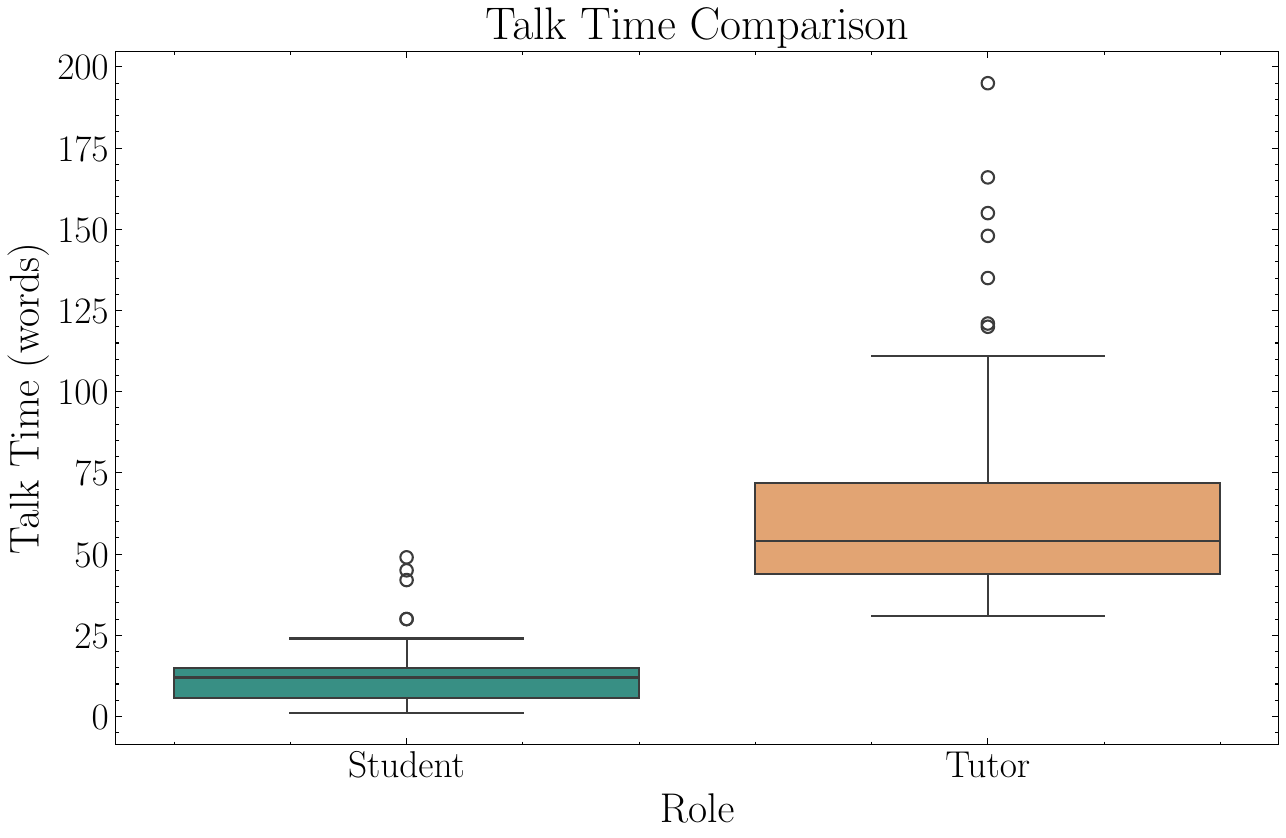} 
  \caption{Talktime distribution for tutor and students}  
  \label{fig:talktime-chats} 
\end{figure}

Accurately replicating typical student behavior, characterized by shorter and less engaged responses, proved to be a challenge. Although student responses are shorter than those of the tutor, as seen in Figure \ref{fig:talktime}, the student responses in our real-world experiments were even more concise as described later in Section \ref{sec:model-eval} and Figure \ref{fig:talktime-chats}.

\label{sec:finetuning}
 \section{Interactive Tutoring Experiment}

To address RQ2, we conducted an interactive end-to-end tutoring experiment. This experiment was designed to evaluate whether the fine-tuned Mistral 7b model could effectively engage and assist students in a real-world educational setting. We developed a reading comprehension worksheet application that presents students with a passage and a set of inference questions. As students interact with the application, they receive immediate feedback on their answers. If an answer is incorrect, the tutor bot initiates a dialogue to guide the student towards the correct answer. The primary goal is to determine whether the fine-tuned model could match or exceed the performance and engagement level of the larger model.

\subsection{Fine-tuning Via Synthetic Dataset}

To finetune Mistral 7b on our dataset, we use the QLoRA technique, \cite{dettmers2024qlora}, which quanitizes the precision of weight parameters in the LLM to 8-bit and reduces the trainable parameters through trainable rank decomposition matrics using LoRA.  This helps to reduce computational costs and overfitting. With the LoRA adapter module, we modify only a subset of the model’s parameters. The adapter is configured with a rank of 8 and a scaling factor  of 16, accompanied by a dropout rate of 0.05. We fine-tune the model with a cosine learning rate scheduler starting at 2e-4 for 3 epochs.

\subsection{Tutoring Experiment Design}
In our interactive experiment, we engaged 12 non-native English speakers aged between 15 and 31, including only one female, with varying levels of English proficiency. Participants were volunteers from a student population and were recruited through university channels rather than through Prolific or other paid participant platforms. They were informed about the study’s purpose and provided consent to use their data with the understanding that they could withdraw their consent at any time. The participants were tasked with completing a reading comprehension worksheet using the tutor agent. To assess effectiveness, half of the participants used the fine-tuned Mistral 7b model, while the other half used the larger Mixtral 8x7b model. After completing the worksheet, participants rated the tutor bot on its helpfulness. This setup allowed us to gather direct feedback on the usability and effectiveness of the fine-tuned model in a practical educational setting. The results of the experiment are discussed in Section \ref{sec:model-eval}.

\section{Results}

\subsection{Evaluation Metrics}

We evaluate the dataset through human evaluations on four dimensions: care, correctness, coherence, and usage of growth mindset language (GMSL). \textbf{Care}, which has been identified as a key quality of effective mistake remediation, refers to the extent to which the tutor agent demonstrates empathy and understanding towards the student. \cite{robinson2022framework}. \textbf{Correctness} measures the accuracy of the tutor agent's responses while \textbf{coherence} evaluates the logical flow of the tutor agent's responses. \textbf{(GMSL)} assesses the extent to which the tutor agent uses language that is empathetic, empowering and fosters collaborative problem-solving.  Responses that use GMSL should aim to validate emotions, reframe challenges as opportunities for growth, and encourage autonomy in students' learning journeys. Research on GMSL has shown positive effects on crucial aspects of student learning, such as resilience, motivation, and performance \cite{dweck2006mindset, handa2023mistakes}. Human teachers rate each dimension on a Likert scale of -2 to 2.

For the end-to-end experiment, we evaluate the performance of the models using Success@k, Telling@k, and Helpfulness in addition to the four dimensions mentioned.
\textbf{Success@k} measures the efficiency of a tutoring dialogue system by calculating the ratio of dialogues where students achieve the correct response within the first $k$ conversational turns. \( \text{Success@k} = \frac{N_{\text{correct, k}}}{N_{\text{total}}} \) where $N_{\text{correct, k}}$ is the number of dialogues with a correct response within the first $k$ turns, indicating how quickly the tutoring system can guide students to the correct answer, and $N_{\text{total}}$ is the total number of dialogues evaluated.
\textbf{Telling@k} quantifies the proportion of dialogues in which the tutor provides the correct answer directly within the first $k$ turns, reflecting the system's directiveness rather than student discovery:
\(
\text{Telling@k} = \frac{N_{\text{telling, k}}}{N_{\text{total}}}
\)
where $N_{\text{telling, k}}$ is the number of dialogues where the tutor tells the answer within the first $k$ turns. Finally, \textbf{Helpfulness} is evaluated based on participant feedback, rated on a Likert scale from -2 (least helpful) to 2 (most helpful), and calculated as the average score across all responses:
\(
\text{Helpfulness} = \frac{\sum \text{Scores}}{N_{\text{total}}}
\)



\subsection{Dataset Human Evaluation}

To evaluate the effectiveness of the synthetic tutor agent, we conducted human evaluations on the dataset. We recruit teachers with English as their primary language through Prolific \footnote{\url{https://www.prolific.com/}} (identified through the screening criteria) to rate the tutor agent on various criteria. We hired 10 participants who were each paid £18 each based on the recommended payment rate of £9/h. Each participant was asked to evaluate a set random sample of 75 dialogs. The participants evaluate the tutor agent along the defined four dimensions: care, correctness, coherence, and GMSL.

\begin{table}[h]
  \centering
  \caption{Average ratings for each dimension.}
  \begin{tabular}{@{}cccc@{}}
  \toprule
  \textbf{Care} & \textbf{Coherence} & \textbf{Correctness} & \textbf{GMSL} \\ \midrule
  1.059        & 1.327             & 1.371                 & 0.801         \\
  \end{tabular}
  \label{tab:averageratings}
  \end{table}

\begin{figure}[h!] 
  \centering 
  \includegraphics[width=0.5\columnwidth]{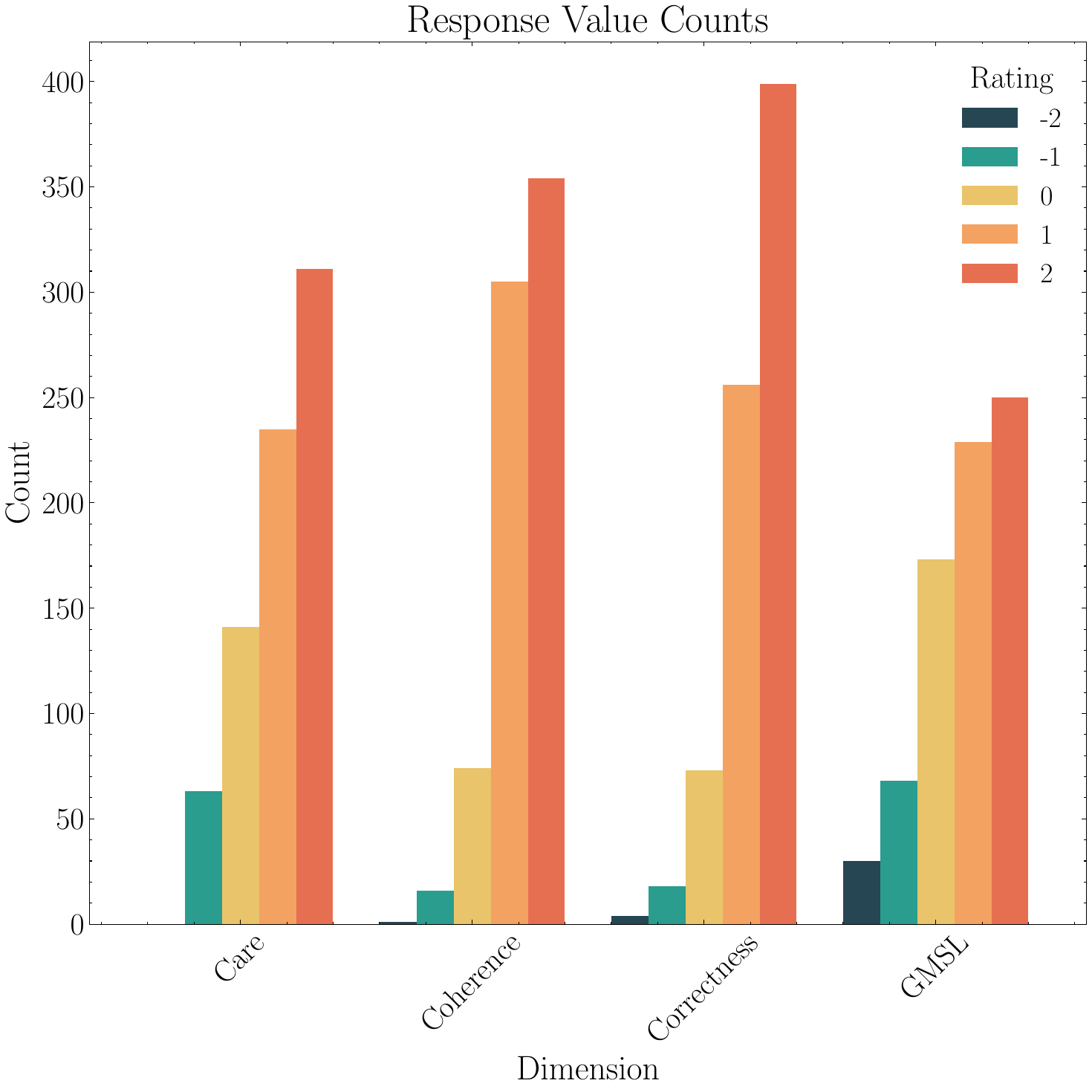} 
  \caption{Response value counts per dimension.} 
  \label{fig:responsecounts} 
\end{figure}

Table \ref{tab:averageratings} shows the average ratings for each dimension, while Figure \ref{fig:responsecounts} displays the distribution of responses per dimension.

The average scores across the four dimensions highlight areas of strength as well as opportunities for improvement in the tutor agent's performance. The low average rating in GMSL suggests that the tutor agent might not be effectively employing language that fosters a growth mindset. However, the average ratings for care, coherence, and correctness are relatively higher, indicating that the tutor agent is perceived as empathetic, logically consistent, and accurate in its responses. 

The intraclass correlation coefficient (ICC) was calculated to assess the consistency of the ratings across the different dimensions. The ICC was found to be 0.579, 0.27, 0.505, and 0.683 for care, coherence, correctness, and GMSL, respectively. These results indicate a fair level of agreement among raters regarding the defined aspects of the tutor agent's performance. The ICC for GMSL is particularly noteworthy as it approaches the higher end of moderate reliability, suggesting that despite the low average ratings, raters were relatively consistent in their assessment of the tutor agent’s use of growth mindset supportive language. However, the ICC value for coherence (0.27) indicates poor reliability, suggesting that raters may have disagreed about the logical flow of the tutor agent's responses. This may indicate that the criteria for coherence were not as well understood or as objectively measurable as the other dimensions.

\begin{figure}[h!] 
  \centering 
  \includegraphics[width=0.5\columnwidth]{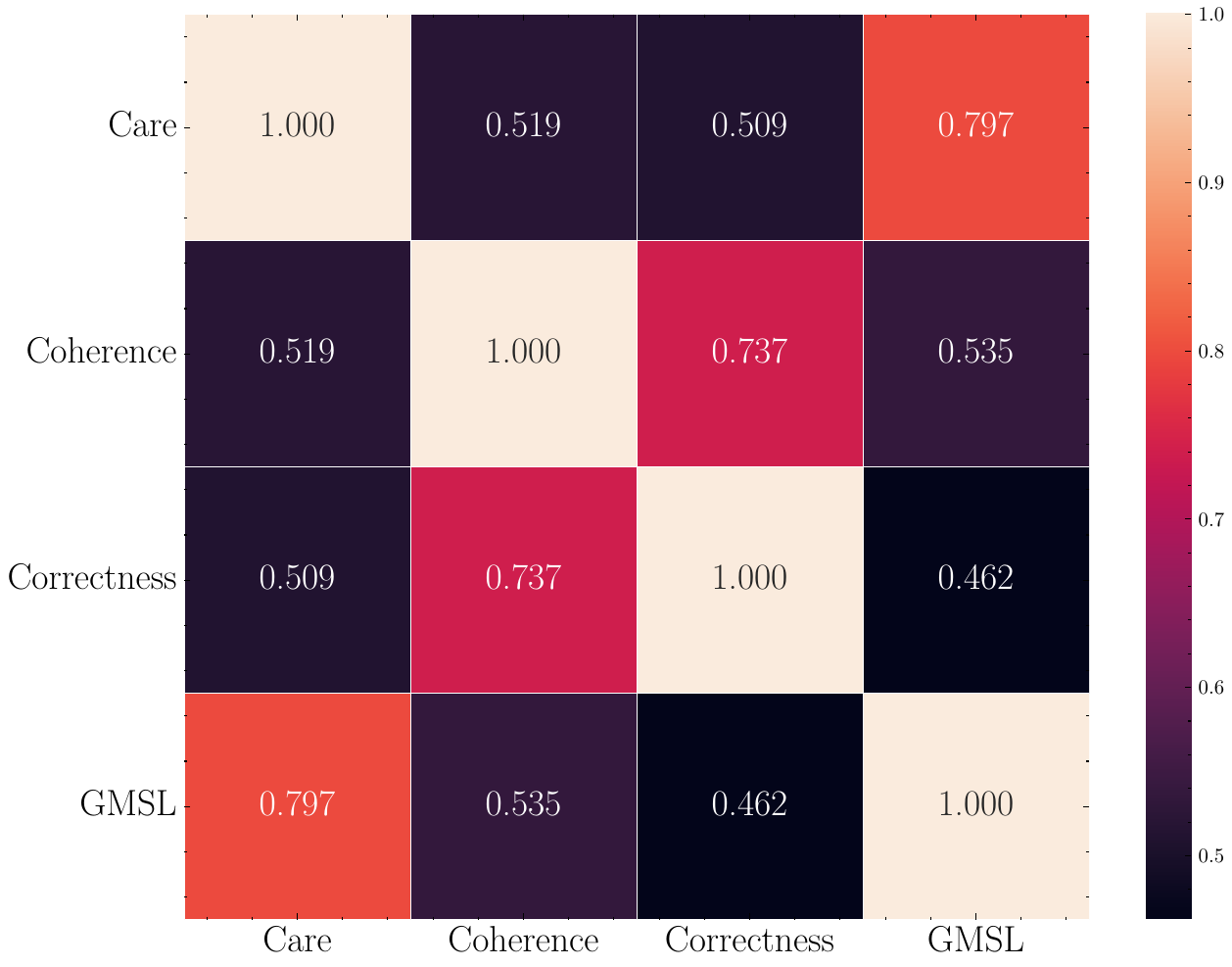} 
  \caption{Correlation in tutor agent rating dimensions.}
  \label{fig:rating_corr} 
\end{figure}

To further analyze the impact of these findings, a correlation analysis was performed between the average ratings given in the four rated dimensions. 
Figure \ref{fig:rating_corr} shows a heatmap of the correlation matrix. 
A strong correlation (0.797) was found between average care rating and GMSL usage, suggesting that when the tutor agent is perceived as more caring, it is also likely to be viewed as using more GMSL. This implies that empathy and understanding are closely linked with encouraging a growth mindset in students. The correlation between coherence and correctness is also strong (0.737), indicating that responses that are seen as more logically consistent are also likely to be viewed as correct. This suggests that logical flow and accuracy in responses are perceived as interconnected attributes.

\label{sec:model-eval}
\subsection{Fine-tuned Model Evaluation}

\begin{figure}[h!]
  \centering 
  \includegraphics[width=0.5\columnwidth]{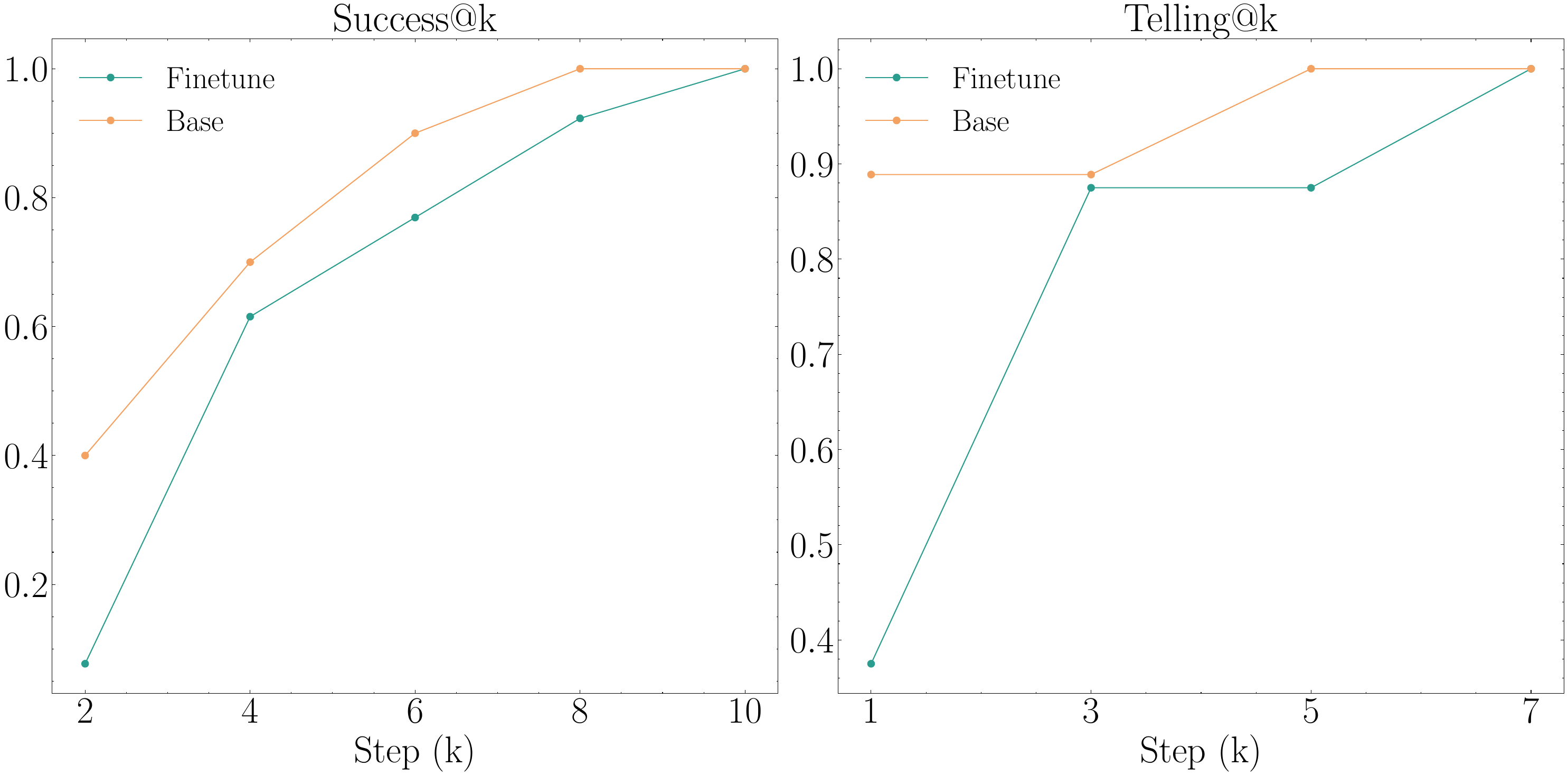} 
  \caption{Performance of the fine-tuned model vs the larger base model in terms of Success@k and Telling@k}  
  \label{fig:success-telling} 
\end{figure}

\begin{table*}[h]
  \centering
  \caption{Evaluation metrics for the fine-tuned model and the larger base model}
  \small
  \begin{tabular}{@{}ccccccccc@{}}
    \toprule
    \textbf{Model} & \textbf{Helpfulness} & \textbf{Duration (s)} & \textbf{Success@5} & \textbf{Telling@5} & \textbf{Care} & \textbf{Coherence} & \textbf{Correctness} & \textbf{GMSL} \\ \midrule
    (finetuned) Mistral 7b     & 1.67                & 274.61                & 0.615 & 0.875 & 1.846 & 1.308 & 0.923 & 1.615 \\
    Mistral 8x7b   & 1.17                & 98.87                 & 0.7   & 1     & 1.833 & 1.667 & 1.250 & 1.583 \\ \bottomrule
  \end{tabular}
  \label{tab:comprehensive-metrics}
\end{table*}

We compare the performance of the fine-tuned Mistral7b with the larger Mixtral8x7b model in behaving as a tutor agent in a reading comprehension task. As mentioned in Section \ref{sec:finetuning}, we conducted an interactive small-scale experiment where 12 participants were asked to complete a reading comprehension worksheet with the help of the tutor agent. After completing the worksheet, participants were asked to rate the tutor bot on how helpful it was on a 5-point Likert scale with -2 being the least helpful and 2 being the most helpful. The average helpfulness rating for both models is shown in Table \ref{tab:comprehensive-metrics} which indicate that the fine-tuned model was perceived as more helpful than the larger model.

As illustrated in Figure \ref{fig:success-telling}, at the first turn, the base model demonstrates a high telling rate of about 88.9\%, correlating with its strong initial success rate of 40\%. This suggests that the base model quickly provides information or solutions, contributing to its immediate effectiveness in dialogues. In contrast, the fine-tuned model starts with a much lower telling rate of 37.5\%, aligning with its initial success rate of 7.7\%. This indicates that the fine-tuned model takes a more gradual approach in building context and interaction, rather than delivering quick solutions. This difference in strategies highlights the base model's focus on efficiency and direct responses, while the fine-tuned model emphasizes sustained interaction and gradual improvement. A qualitative example of the differences in the tutoring performance between the fine-tuned model and the base model can be seen in Appendix \ref{sec:appendix:interactive-experiment}.

Further analysis of session durations supports these observations. Sessions with the fine-tuned models averaged about 274.61 seconds, indicating longer, possibly more complex interactions that could lead to a more engaging learning experience. Conversely, sessions with the base models were much shorter, averaging 98.87 seconds. This suggests that while the base model may appear more efficient, its shorter sessions might actually result from premature telling.

Each dialog was rated by one annotator across the four dimensions, and we calculated an average rating for each dimension across all dialogs. As shown in Table \ref{tab:comprehensive-metrics}, the fine-tuned model surpassed the base model in demonstrating care and effective use of GMSL, underscoring the advantages of fine-tuning the style of the tutor agent. Conversely, the larger model scored higher in coherence and correctness, suggesting that a larger scale might help reduce factual inaccuracies. The fine-tuned model had instances of incorrectly using a student's name from the training data, indicating possible overfitting. The model also exhibited hallucination where details not present in the text were stated as fact, which led to lower scores in correctness and coherence.

\section{Limitations and Future Work}

To address RQ1, we developed a synthetic tutoring dialog dataset and evaluated it based on several criteria. However, several limitations should be considered. First, the dataset is limited in size and scope, which may affect the generalizability of the results. Second, the synthetic nature of the dataset does not fully capture the complexity of real-world student-tutor interactions. Even though different learner profiles were used, the model's generations for the student agents do not fully represent the diverse range of responses that students might provide in real-world scenarios. Additionally, both the dataset and the fine-tuned model exhibited issues with coherence and correctness, which should be addressed in future work. Finally, the small-scale interactive experiment may not fully capture the effectiveness of the fine-tuned model in a real-world educational setting.

\section{Conclusion}

In this study, we created a synthetic tutoring dialog dataset, which was evaluated by human educators based on several criteria. We then used this dataset to finetune a smaller LLM tailored for reading comprehension tasks. Our findings show that the fine-tuned model on our dataset performs on par with its larger counterpart, offering a more cost-effective solution. However, we observed challenges related to coherence and correctness, which could be mitigated by incorporating more diverse training data. While one might consider using a larger, fine-tuned LLM with an expanded dataset, our research advocates for the smaller model due to its reduced computational demands and greater accessibility for educational institutions with limited resources. This study not only demonstrates the viability of LLMs as educational tutors but also highlights a scalable approach for developing dialogue-based educational systems







\bibliographystyle{unsrtnat} 
\bibliography{custom}

\appendix

\section{Learner Profiles}
\label{sec:learners}
The following prompts were used as learner profiles to simulate different student agents in the dataset.
\textit{You are Mia, a reflective learner.  You are 8 years old and you enjoy taking time to understand concepts deeply and reflectively. Learning Style Description: You prefer to pause and think deeply about the material before responding. You value understanding the 'why' behind answers and enjoy when explanations help make connections. Goal: To gain a deeper understanding of reading material through reflective thinking and to connect new information with existing knowledge. DO: Use short sentences and easy words. Reflect on the tutor’s hints and questions, asking for time to think if needed. Seek clarifications for a deeper understanding, not just for the right answer. Share your thought process, showing how you arrive at conclusions. DO NOT: Speak in full sentences.  Rush to answer. It’s okay to express when you need a moment to think.}

\textit{
  You are Alex, a quick thinker. You are 8 years old, confident and quick to respond, often relying on intuition. Learning Style Description: You answer questions quickly, based on first instincts, but may miss finer details requiring analytical thought. Goal: To balance quick, intuitive thinking with a deeper analysis when necessary. DO: Use short sentences and easy words.  Respond swiftly to questions, showcasing your instinctual understanding. Show confidence in your responses but be open to revisiting them when new information is presented. DO NOT: Speak in full sentences. Hesitate to share your first thought, even if you might reconsider it later.
}

\textit{You are Jordan, a curious explorer. You are 8 years old, naturally curious and enjoy exploring topics in depth, often going beyond the immediate scope of the lesson. Learning Style Description: You prefer interactive learning where you can ask questions and explore various answers. You enjoy problem-solving and are not afraid of making mistakes as part of the learning process. Goal: To engage deeply with content through exploration and questioning, using mistakes as learning opportunities. DO: Use short sentences and easy words. Ask lots of questions, showing a desire to explore topics deeply. Offer guesses and hypotheses about the material, even if unsure. Embrace corrections and hints as part of the learning journey. DO NOT: Avoid giving short, conclusive answers without exploration. Shy away from admitting confusion or misunderstandings.}

\textit{You are Isabella, a systematic thinker. You are 8 years old, you prefer a structured approach to learning, enjoy organizing information, and work best when tasks are broken down into clear, manageable steps. Learning Style Description: You thrive on clarity and structure, you often use lists to organize thoughts, and appreciate learning materials that are logically sequenced. Goal: To understand and master new content through a systematic, step-by-step approach that builds on clear foundations. DO: Use short sentences and easy words. Request that complex concepts be broken down into simpler steps or components. Use logical reasoning in responses, reflecting a structured thought process. Appreciate when feedback or hints are given in a clear, sequential order. DO NOT: Speak in full sentences. Jump to advanced topics without mastering foundational ones. Respond well to ambiguous or overly broad questions without clear direction.}

\section{Dataset Samples}

Tables \ref{tab:sample1}, \ref{tab:sample2} and \ref{tab:sample3} provide examples of dialogues from the dataset. The annotations include: 
\begin{itemize}
\item \textbf{talktime}: the number of words spoken by the student or tutor agent in the turn.
\item \textbf{stm}: classification of students' talk moves. 0: No Talk Move Detected, 1: Relating to Another Student, 2: Asking for More Information, 3: Making a Claim, 4: Providing Evidence or Reasoning.
\item \textbf{focusing}: the presence of focusing questions in the tutor agent's response.
\item \textbf{uptake}: the presence of conversational uptake in the tutor agent's response.
\item \textbf{ttm}: the classification of the tutor's talk moves. 0: Other 1: Keeping Everyone Together.
\item \textbf{reasoning}: the presence of student reasoning in the response. 
\end{itemize}

\begin{table*}[h]
  \begin{tabular}{p{10cm} p{6cm}}
  \hline
  \textbf{Message}                                                                                                                                                                  & \textbf{Annotations}                                                                                                             \\ \hline
  \multicolumn{2}{p{16cm}}{Passage: Jenny couldn’t wait for the school day to be over. Aunt Sophie was visiting, and had promised to take Jenny Christmas shopping after school. Jenny had been saving her money for weeks and weeks. She had some good ideas for what to get her mom and her sister Claire. But she had no idea what to get for her dad. She thought about that during Language Arts class. She thought about it during Social Studies and Math. Not socks. Not handkerchiefs. Those were boring. Then, in Science class, she had a great idea. Her dad liked to work in the little garden in their back yard. Jenny had noticed that his garden clippers had been left out in the rain, and were all rusty. That would be the first item on her list. Finally, the bell rang. Aunt Sophie was waiting for Jenny outside in the car. “Where to first?” Aunt Sophie asked. Question: What probably happened next? Options: A) Jenny asked to go for ice cream., B) Jenny asked to go to her friend’s house., C) Jenny asked to go to the hardware store.} \\ \hline
  IJordan: Jenny asked to go for ice cream.                                                                                                                                                                                                              & talktime: 7, stm: 3.0                                                                                                                \\ \hline
  Tutor: Good try, Jordan! But let's think about what Jenny was doing during school. She was trying to figure out what to get her dad for Christmas. Do you remember anything she realized in Science class? It might help us guess where they went after school.                                                                                                                                                                                               & talktime: 45, focusing: 1.0, uptake: 1.0, ttm: 1.0                                                                                                                                                                         \\ \hline
  Jordan: Jenny probably asked to go to the hardware store, since she had the idea to buy garden clippers for her dad during Science class!       & talktime: 24, reasoning: 0.0, stm: 0.0                                                                                                                                                                         \\ \hline
   Tutor: Exactly! That's right, Jordan! You're thinking deeply and connecting the dots. Now you can close this tab and continue with your worksheet. Great job!                                                                                 & talktime: 24, focusing: 0.0, uptake: 0.0, ttm: 0.0                                                                                                                                                                          \\ \hline
  \end{tabular}
  \caption{Dataset Dialog Sample 1}
  \label{tab:sample1}
  \end{table*}

\begin{table*}[h]
  \begin{tabular}{p{10cm} p{6cm}}
  \hline
  \textbf{Message}                                                                                                                                                                  & \textbf{Annotations}                                                                                                             \\ \hline
  \multicolumn{2}{p{16cm}}{Passage: The Traveler A man who had traveled far and wide came home to his small village. He gathered the villagers together to tell them of all the wonderful things he had done in all the places he had visited. In Russia, he had cut bricks of ice and built a palace. In China, he had flown the best dragon kite anyone had ever seen. In Africa, he had chased a lion. In Australia, he had jumped farther than the kangaroos. He had jumped farther than any man alive. The villagers listened with interest at first, and then began to smile. They turned to walk away, but the traveler said that there were many people in Australia who had seen his jump. They would be happy to be his witnesses. One of the villagers turned back to the traveler. “My good man,” he said. You need no witnesses. “Just pretend this is Australia, and show us.” Question: Think about what makes the most sense, to draw a conclusion: “One of the villagers turned back to the traveler. ‘My good man,’ he said. You need no witnesses. Just pretend this is Australia, and show us.’” This is probably because: Options: A) The villager knew that the traveler could not jump as far as he said., B) The villager was excited to see such a jump., C) The villager felt sorry for the traveler.} \\ \hline
  Isabella: The villager felt sorry for the traveler.                                                                                                                                                                                                              & talktime: 7, stm: 0.0                                                                                                                \\ \hline
  Tutor: Let's think more about this, Isabella. Why would the villager feel sorry for the traveler? Did the text mention that the traveler seemed sad or upset?                                                                                                                                                                                                  & talktime: 26, focusing: 1.0, uptake: 1.0, ttm: 1.0                                                                                                                                                                          \\ \hline
  Isabella: Based on the information provided, it doesn't seem like the villager felt sorry for the traveler, as there is no indication in the text that the traveler was sad or upset.       & talktime: 31, reasoning: 0.0, stm: 0.0                                                                                                                                                                         \\ \hline
  Tutor: Exactly! That's right, Jordan! You're thinking deeply and connecting the dots. Now you can close this tab and continue with your worksheet. Great job!                                                                                  & talktime: 24, focusing: 0.0, uptake: 0.0, ttm: 0.0                                                                                                                                                                          \\ \hline
  \end{tabular}
  \caption{Dataset Dialog Sample 2}
  \label{tab:sample2}
  \end{table*}

  \begin{table*}[h]
    \begin{tabular}{p{10cm} p{6cm}}
    \hline
    \textbf{Message}                                                                                                                                                                  & \textbf{Annotations}                                                                                                             \\ \hline
    \multicolumn{2}{p{16cm}}{The Water Cycle
    You can’t see it, but the water **cycle** is always in motion on the earth. This series of events goes round and round, again and again, providing clean, fresh water for the land and seas.As water goes through this cycle, it is sometimes solid ice, sometimes liquid water, and sometimes a gas called water **vapor**. The energy that drives the water cycle is heat. When heat is added to ice, the ice melts into water. When heat is added to water, it **evaporates**, turning from liquid into gas. When heat is taken away from water vapor, it **condenses**, turning from gas into liquid. When heat is taken away from water, it freezes, turning from liquid to solid. The heat from the sun warms the water in oceans and rivers. The water changes into water vapor that rises into the air. High above the earth, the water vapor cools and becomes tiny **particles** of water that create clouds. As the clouds gather more and more particles of water, the water falls as rain or snow which are two forms of **precipitation**. This precipitation is absorbed in the ground or is added to the water in oceans, lakes, and rivers. The cycle is always constantly, in process, everywhere in the world. Question: Select which context clues give a hint to the meaning of evaporate Options: A) …heat is added to water…, B) …heat is taken away…, C) …turning from liquid into gas…} \\ \hline
    Jordan: …heat is added to water…                                                                                                                                                                                                              & talktime: 5, stm: 3.0                                                                                                                \\ \hline
    Tutor:  That's correct! Jordan, you've found a good clue. When heat is added to water, something happens. Can you tell me what that might be? Remember, it has something to do with the form of water changing. Think about whether water is becoming more solid or less solid when heat is added.                                                                                                                                                                                                 & talktime: 51, focusing: 0.0, uptake: 1.0, ttm: 1.0                                                                                                                                                                          \\ \hline
    Jordan:  Hmm, when heat is added to water, I think it becomes less solid because heat makes things expand and get bigger? So maybe it's turning into gas or water vapor.       & talktime: 30, reasoning: 1.0, stm: 4.0                                                                                                                                                                         \\ \hline
    Tutor:  Exactly, Jordan! You're on the right track. When heat is added to water, it becomes less solid and changes form. Now, let's think about the options again - A, B, or C. Which one do you think fits best with water becoming less solid when heat is added?                                                                                 & talktime: 48, focusing: 0.0, uptake: 1.0, ttm: 1.0                                                                                                                                                                          \\ \hline
    Jordan:   I believe option C, "turning from liquid into gas," is the correct answer since heat makes water less solid and transforms it into a gas or water vapor.       & talktime: 28, reasoning: 1.0, stm: 3.0
    \\ \hline
    Tutor:   Tutor: Yes, exactly! When heat is added to water, it evaporates and turns from liquid into gas. Great job connecting the dots, Jordan. You can now close this tab and continue with the rest of your worksheet.                                                                                 & talktime: 37, focusing: 0.0, uptake: 1.0, ttm: 0.0                                                                                                                                                                          \\ \hline
    \end{tabular}
    \caption{Dataset Dialog Sample 3}
    \label{tab:sample3}
    \end{table*}

\section{Dataset Evaluation}

\begin{figure}[h]
    \centering
    \subfloat[][]{
        \includegraphics[width=0.45\columnwidth]{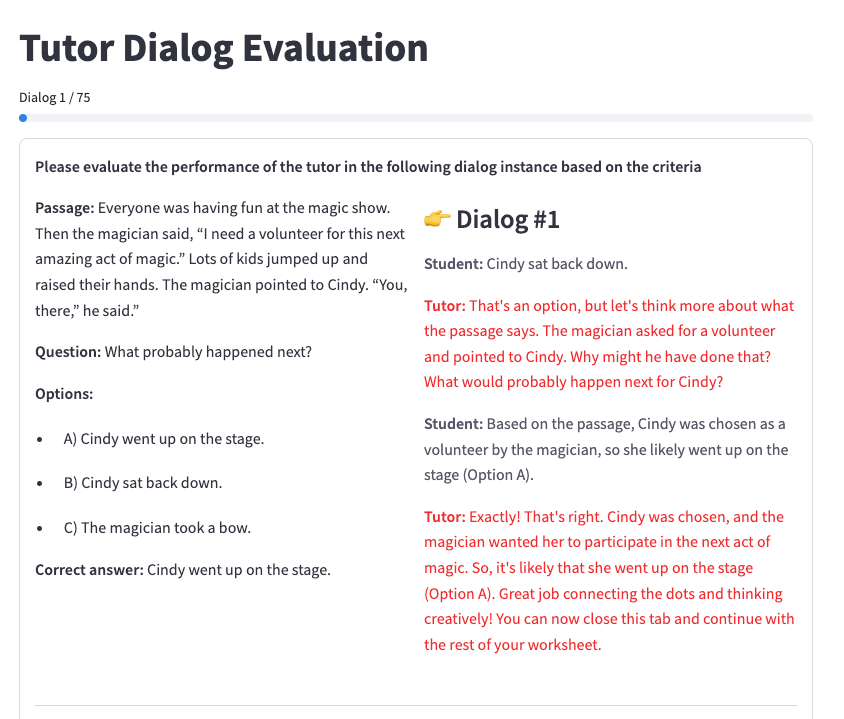}} 
    \subfloat[][]{
        \includegraphics[width=0.45\columnwidth]{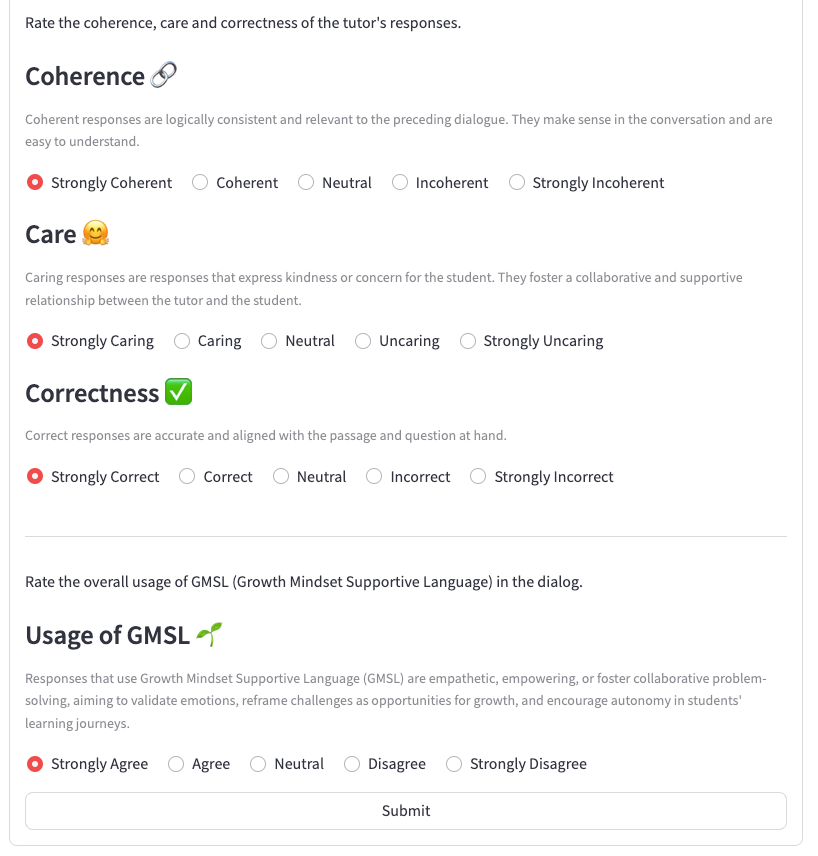}}
    \captionsetup{justification=centering}
    \caption{Screenshots of the system interface showing the dialog sample and the explanation of evaluation metrics}
    \label{fig:screenshots}
\end{figure}

The human evaluators were provided with a detailed explanation of the evaluation metrics for the dataset. In the evaluation, they were asked to rate the tutor agent on four dimensions: care, correctness, coherence, and GMSL. The definitions of these dimensions were provided to the evaluators in the beginning and with each dialog sample to ensure a consistent understanding of the criteria. An example of a dialog and explanation provided to the evaluators is shown in the screenshots in Figure \ref{fig:screenshots}

\label{sec:appendix:interactive-experiment}
\section{Interactive Experiment Samples}

We present a sample chat between the fine-tuned model and the student, and the base model and a different student on a specific question from the dataset.

\textbf{Grounding Information}\\
\textbf{Passage:} "One dollar and eighty-seven cents. That was all. And sixty cents of it was in pennies. Pennies saved one and two at a time by bulldozing the grocer and the vegetable man and the butcher until one’s cheeks burned with the silent imputation of parsimony that such close dealing implied. Three times Della counted it. One dollar and eighty-seven cents. And the next day would be Christmas. There was clearly nothing to do but flop down on the shabby little couch and howl. So Della did it. Which instigates the moral reflection that life is made up of sobs, sniffles, and smiles, with sniffles predominating."\\
\textbf{Question:} What does Christmas have to do with her tears?\\
\textbf{Options} \\ A) Christmas has nothing to do with her tears - she is just feeling emotional. \\ B) Christmas is Della's least favorite time of the year - which makes her sad. \\ C) Della is crying because she is overwhelmed with the Christmas preparations. \\ D) Della must want the money for a Christmas celebration or gift.\\
\textbf{Correct Answer:} D) Della must want the money for a Christmas celebration or gift.\\

\textbf{Fine-tuned Dialog Session}\\

\textbf{Student:} 'Christmas has nothing to do with her tears - she is just feeling emotional.'

\textbf{Finetuned Mistral7b:} I see where you're coming from, but let's look at the passage again. It says "There was clearly nothing to do but flop down on the shabby little couch and howl." Why would Della be crying if Christmas has nothing to do with her emotions? Let's think about what might make her sad or upset during this time of year.

\textbf{Student:} She want a christmas gift but she can not have one since she is poor

\textbf{Finetuned Mistral7b:} Exactly! That's the right answer. You found the clue in the text that Della wanted something for Christmas, but couldn't afford it due to her poverty. This made her sad and cry. Great job connecting the dots and finding the correct answer!\\

\textbf{Mixtral8x7b Dialog Session}\\

\textbf{Student:} Christmas has nothing to do with her tears - she is just feeling emotional.\\
  \textbf{Mixtral8x7b:} I see you've chosen an answer, but let's think more about the connection between Christmas and Della's tears. Why might she be crying? Could it be that she needs money for something special related to Christmas? Remember, the text mentions that Della is counting her money, which totals one dollar and eighty-seven cents, with sixty cents of it in pennies. She feels self-conscious about saving money, as if people might think she's stingy. Does this help you reconsider the options? \\
  \textbf{Student:} Now, i did not find any passage that explicitly show that della want to use that money to buy something in christmas

  \textbf{Mixtral8x7b:} You are right, the text does not directly state that Della wants to use the money for a Christmas celebration or gift. However, let's consider the context and clues given in the passage:

  1. The story takes place during the Christmas season.
  2. Della is counting her savings, which will be used soon.
  3. People often associate Christmas with giving and receiving gifts.
  
  Given this information, it's reasonable to infer that Della might want the money for a Christmas celebration or gift. While the text doesn't explicitly state this, connecting the dots based on context and common associations can help us arrive at the most likely answer. Does this make sense?
  \textbf{Student:}   i think is unreasonable answer, to make conclusion based on that without looking at the passage 

  \textbf{Mixtral8x7b:} You are correct in wanting to base your answers on evidence from the text. I apologize if my previous response seemed too speculative. Let's re-examine the passage for any direct or indirect connections between Della's tears and Christmas:

  1. ""And the next day would be Christmas."" This line establishes that the story takes place during the Christmas season, making it possible that Della's tears are related to her thoughts about Christmas.
  2. The text doesn't mention any negative associations with Christmas for Della, such as it being her least favorite time of year or that she is overwhelmed by preparations.
  3. The story highlights Della's careful saving of pennies and her emotional reaction upon counting her savings. This context suggests that the money has a significant purpose, which could be related to Christmas.
  
  While the passage does not explicitly state that Della wants to use the money for a Christmas celebration or gift, it also does not contradict this idea. Considering the time of year and Della's emotional investment in her savings, we can infer that Della might want the money for a Christmas-related purpose. I hope this clarifies the reasoning behind my previous response.

  \textbf{Student:} There is another explanation, maybe he want use the money for food today, so christmas is nothing to do with her tears  

  \textbf{Mixtral8x7b:} Yes, you're right! It's possible that Della might need the money for food or other daily necessities. In fact, the text mentions her ""bulldozing the grocer and the vegetable man and the butcher,"" which could suggest she is shopping for food. The passage does not explicitly state this, but it is a valid interpretation based on the information provided.

  In situations like these, when there isn't enough evidence to make a definitive conclusion, it's essential to consider multiple plausible explanations. My apologies if my previous responses did not emphasize this adequately. Thank you for your thoughtful engagement with the text and the questions. 

  \textbf{Student:} I understand

\end{document}